\documentclass{article}

\usepackage[preprint, nonatbib]{neurips_2024}

\usepackage[numbers, compress]{natbib}
\usepackage[utf8]{inputenc} 
\usepackage[T1]{fontenc}    
\usepackage{hyperref}       
\usepackage{url}            
\usepackage{booktabs}       
\usepackage{amsfonts}       
\usepackage{nicefrac}       
\usepackage{microtype}      
\usepackage{xcolor}         

\usepackage{amssymb}
\usepackage{amsfonts}
\usepackage{multirow}
\usepackage{tabularx}
\usepackage{graphicx}
\usepackage{amsmath}

\graphicspath{{./figures}}
\DeclareGraphicsExtensions{.pdf,.jpeg,.jpg,.png}

\newcolumntype{V}{>{\raggedright\arraybackslash}X}
\newcolumntype{Y}{>{\centering\arraybackslash}X}
\newcolumntype{Z}{>{\raggedleft\arraybackslash}X}

\newcolumntype{v}{>{\hsize=.5\hsize}V}
\newcolumntype{y}{>{\hsize=.5\hsize}Y}
\newcolumntype{z}{>{\hsize=.5\hsize}Z}

\newcolumntype{o}{>{\hsize=.3\hsize}Z}

\usepackage[capitalize]{cleveref}
\crefname{section}{Sec.}{Secs.}
\Crefname{section}{Section}{Sections}
\Crefname{table}{Table}{Tables}
\crefname{table}{Tab.}{Tabs.}
\title{Fine-Grained Image Recognition \\ from Scratch with \\ Teacher-Guided Data Augmentation}

\author{
    Edwin Arkel Rios\textsuperscript{\dag}, Fernando Mikael\textsuperscript{\ddag}, Oswin Gosal\textsuperscript{\ddag}, Femiloye Oyerinde\textsuperscript{*}, Hao-Chun Liang\textsuperscript{\dag}, \\
    \textbf{Bo-Cheng Lai\textsuperscript{\dag}, Min-Chun Hu\textsuperscript{\ddag} } \\
    \textsuperscript{\dag}\textit{National Yang Ming Chiao Tung University, Taiwan}, \textsuperscript{\ddag}\textit{National Tsing Hua University, Taiwan}, \\
    \textsuperscript{*}\textit{Cohere Labs Community, Nigeria}\\
}

\begin{document}

\maketitle

\begin{abstract}

Fine-grained image recognition (FGIR) aims to distinguish visually similar sub-categories within a broader class, such as identifying bird species. While most existing FGIR methods rely on backbones pretrained on large-scale datasets like ImageNet, this dependence limits adaptability to resource-constrained environments and hinders the development of task-specific architectures tailored to the unique challenges of FGIR.

In this work, we challenge the conventional reliance on pretrained models by demonstrating that high-performance FGIR systems can be trained entirely from scratch. We introduce a novel training framework, TGDA, that integrates data-aware augmentation with weak supervision via a fine-grained-aware teacher model, implemented through knowledge distillation. This framework unlocks the design of task-specific and hardware-aware architectures, including LRNets for low-resolution FGIR and ViTFS, a family of Vision Transformers optimized for efficient inference. 

Extensive experiments across three FGIR benchmarks over diverse settings involving low-resolution and high-resolution inputs show that our method consistently matches or surpasses state-of-the-art pretrained counterparts. In particular, in the low-resolution setting, LRNets trained with TGDA improve accuracy by up to 23\% over prior methods while requiring up to 20.6x less parameters, lower FLOPs, and significantly less training data. Similarly, ViTFS-T can match the performance of a ViT B-16 pretrained on ImageNet-21k while using 15.3x fewer trainable parameters and requiring orders of magnitude less data. These results highlight TGDA's potential as an adaptable alternative to pretraining, paving the way for more efficient fine-grained vision systems.

\end{abstract}

\section{Introduction}
\label{sec:intro}

Fine-grained image recognition (FGIR) focuses on classifying sub-categories within a super-category. Examples include identifying bird species \cite{wah_caltech-ucsd_2011} or car models \cite{krause_3d_2013}. FGIR has diverse applications in multimedia analysis, such as biodiversity monitoring \cite{van_horn_building_2015}, machinery inspection \cite{maji_fine-grained_2013}, and agriculture \cite{wang_identification_2021}.

A key challenge in FGIR is the large intra-class variance and small inter-class variance, which make distinguishing between similar categories difficult \cite{wei_fine-grained_2022}. To address this, numerous methods have been proposed \cite{fu_look_2017, zheng_learning_2017, hu_see_2019, rao_counterfactual_2021}, leading to significant improvements over the past decade. These approaches typically rely on a feature extraction backbone pretrained on large-scale datasets like ImageNet and fine-tuned to capture subtle, discriminative features that differentiate classes.

While these methods primarily differ in how they train and utilize backbones for feature selection, recent studies \cite{ye_image_2024} highlight two key findings: (1) the choice of backbone significantly impacts FGIR performance, and (2) improvements in accuracy are largely driven by backbone size \cite{thompson_computational_2023}. Not only certain backbones are more suitable for certain tasks, but the backbone imposes constraints for training and deployment costs.

These factors motivate us to explore methods for directly training computationally efficient yet effective backbones for FGIR, without relying on pretraining. One promising approach is knowledge distillation (KD).

Knowledge distillation (KD) has emerged as a powerful tool for model compression and transfer learning, in which a high-capacity teacher model guides the training of a smaller student model \cite{hinton_distilling_2015, gou_knowledge_2021}. Previous work applying KD to FGIR settings \cite{zhang_cekdcross_2022, liang_dynamic_2024, zhang_data_2025} have focused on improving accuracy by incorporating data-aware augmentation based on class-activation maps (CAMs) \cite{zhou_learning_2016}. However, these approaches rely on \textbf{one-stage co-training} setups where both teacher and student share the same pretrained backbone. This design assumes the availability of pretrained weights, and limits applicability when training models from scratch, where the student is not yet capable of providing meaningful learning signals to the teacher.

Moreover, CAM-based augmentations are static and limited in diversity, failing to fully capture the rich part-based variations that are crucial in fine-grained tasks. To overcome these limitations, we propose \textbf{T}eacher-\textbf{G}uided \textbf{D}ata \textbf{A}ugmentation (TGDA), a novel \textbf{two-stage} training framework designed to enable training from scratch while retaining the benefits of distillation. TGDA employs a fine-grained-aware fine-tuned teacher model to extract high-quality part attention maps (PAMs), which are used to drive diverse augmentations. These augmented views provide rich supervision for training a student model from random initialization, without any reliance on external pretrained weights.

TGDA’s ability to enable training from scratch opens new possibilities for task-specific architecture design, which is particularly valuable in settings where standard pretrained backbones are suboptimal. One such setting is low-resolution (LR) fine-grained recognition, where input images lack the detail necessary for models to localize subtle discriminative features. This is a common constraint in real-world deployments, where high-resolution imagery is often unavailable. While prior work has attempted to mitigate this challenge by distilling information from high-resolution images \cite{liang_dynamic_2024, zhao_decoupled_2022}, they do not address the inherent architectural limitations of existing backbones.

\begin{figure}[!htb]
    \centering
    \begin{minipage}{0.48\linewidth}
        \centering
        \includegraphics[width=\linewidth]{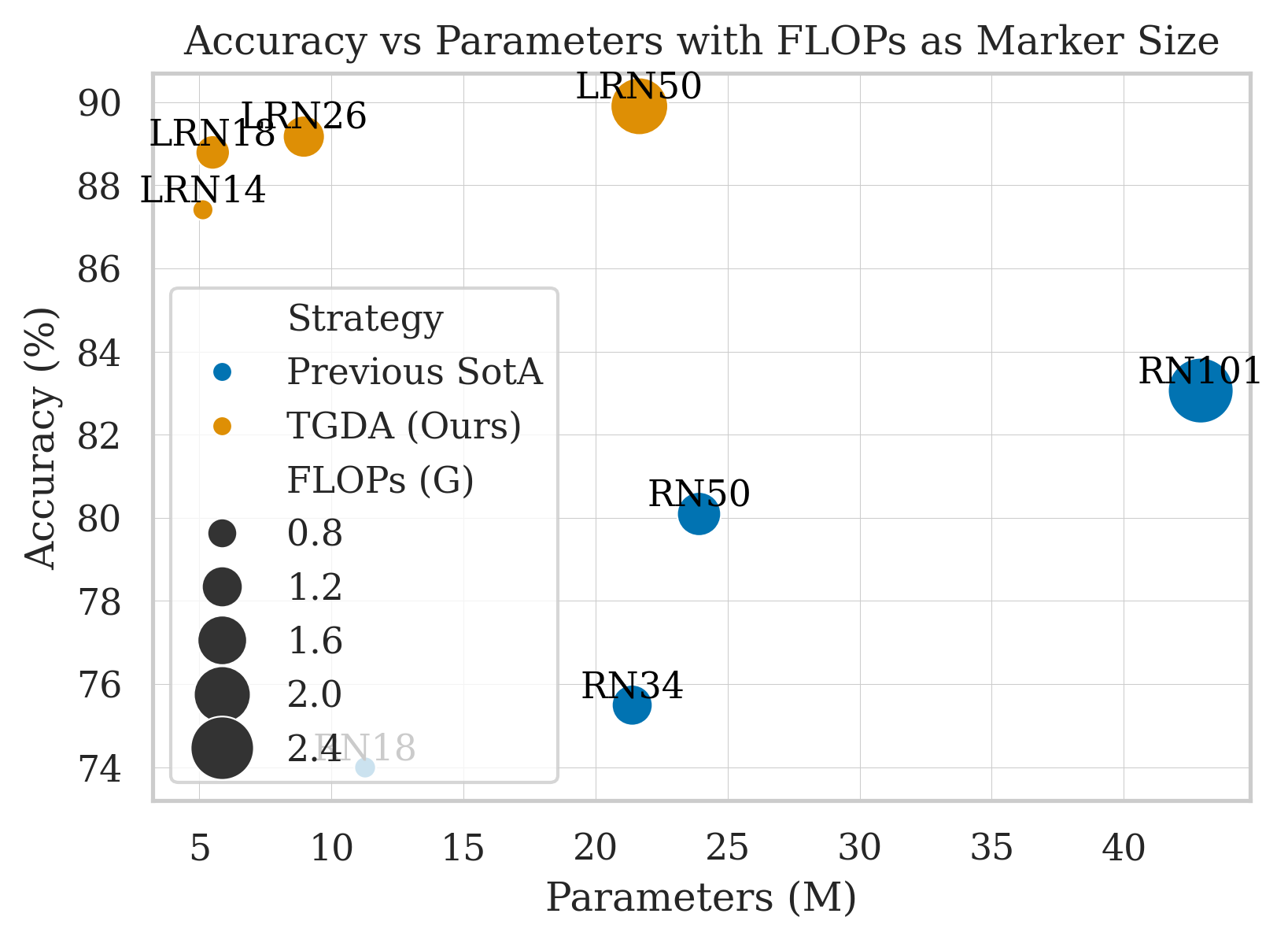}
        \label{figure_acc_vs_cost_is128}
    \end{minipage}
    \hspace{0.01\linewidth} 
    \begin{minipage}{0.48\linewidth}
        \centering
        \includegraphics[width=\linewidth]{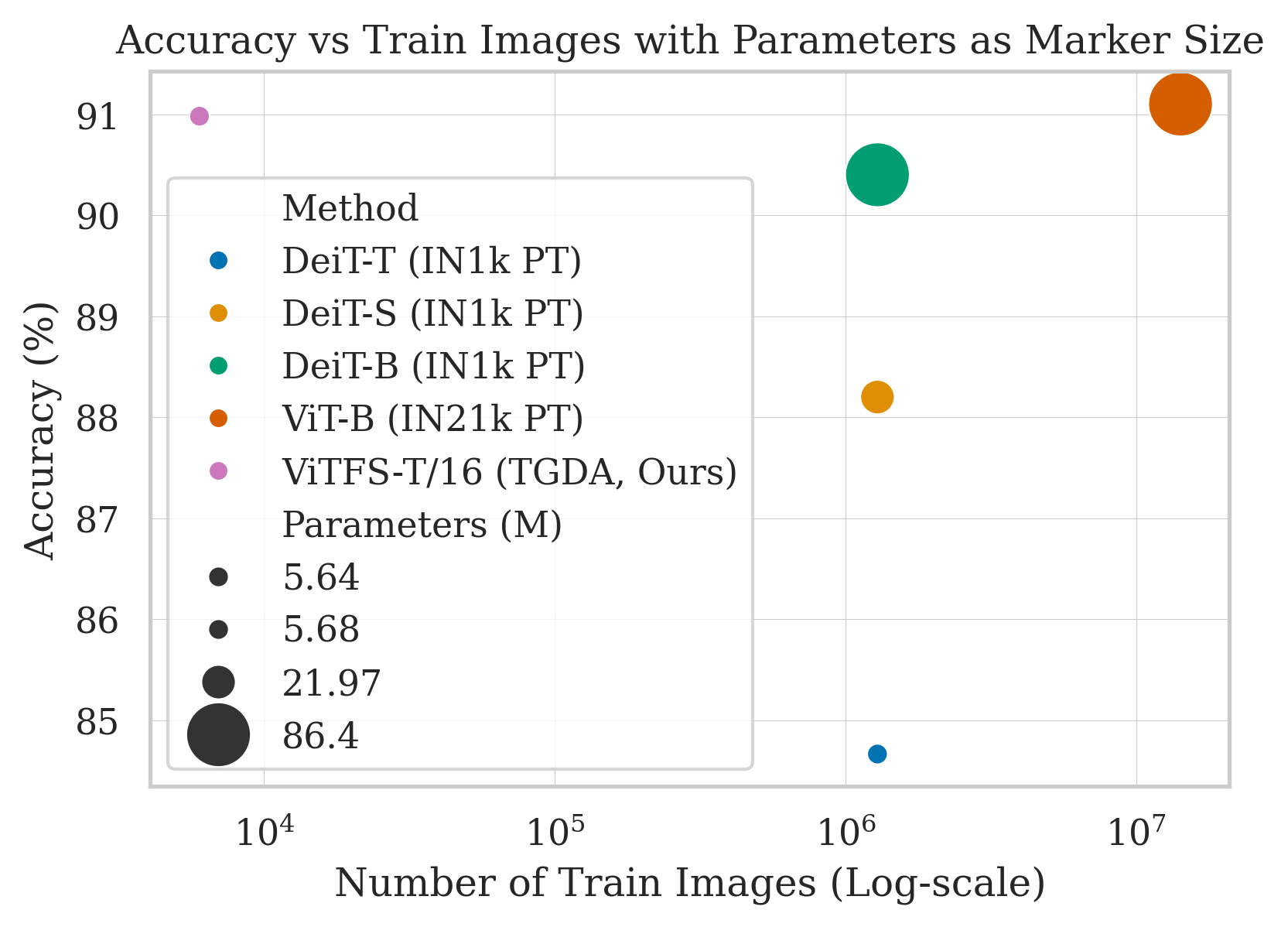}
        \label{figure_acc_vs_cost_vit}
    \end{minipage}
    \caption{Average vs cost trade-offs of our method compared to prior work across diverse FGIR settings. Left: average accuracy across three FGIR datasets (y-axis) vs trainable parameters (x-axis) at low-resolution ($128\times 128$). Marker size indicates number of floating-point-operations (FLOPs). Right: average accuracy (y-axis) vs number of images in the training data (x-axis, log-scale)  ViT-based models using high-resolution inputs (448×448). Marker size indicates number of parameters. Despite being trained on several orders of magnitudes less data, our method can consistently achieve competitive or even superior accuracy with reduced computational cost across diverse FGIR tasks.}
    \label{figure_acc_vs_cost_highlights}

\end{figure}

To this end, we introduce LRNets, a new family of ResNet-based architectures explicitly designed for low-resolution FGIR. By modifying early-stage downsampling operations, LRNets preserve critical fine-grained information and, when trained under TGDA, significantly outperform prior low-resolution methods while requiring less computational resources in terms of model parameters and floating-point-operations (FLOPs) as observed in the left part of \cref {figure_acc_vs_cost_highlights}.

TGDA also facilitates the design of hardware-aware architectures. In particular, vision transformers (ViT) heavily rely on Layer Normalization \cite{ba_layer_2016}, which, while effective during training, introduces inefficiencies and latency during inference for devices optimized for Batch Normalization (BN) \cite{ioffe_batch_2015, noauthor_ethos-u65_nodate}. This mismatch limits the deployability of ViTs in real-world, resource-constrained environments.

To address this, we introduce ViTFS, a family of ViTs trained entirely from scratch under TGDA, featuring BatchNorm and architectural refinements that improve training under limited data. Remarkably, ViTFS-Tiny outperforms a pretrained DeiT-B \cite{touvron_training_2021} model (trained on ImageNet-1k) and matches the accuracy of ViT-B trained on ImageNet-21k while using 15.3× fewer trainable parameters and being trained on $10^2$ to $10^3$ less data, respectively, as shown in the right side of \cref{figure_acc_vs_cost_highlights}. These results highlight TGDA’s ability to replace scale and pretraining with task-specific design, unlocking the full potential of compact, efficient ViTs in fine-grained settings.

In summary, our main contributions are:
\begin{itemize}
    \item A new training framework for FGIR from scratch: We propose Teacher-Guided Data Augmentation (TGDA), a two-stage knowledge distillation framework that enables effective training of student models from random initialization. TGDA leverages a fine-grained-aware teacher to drive diverse, targeted augmentations to eliminate the need for large-scale pretraining while preserving competitive accuracy.
    
    \item Task-specific CNN backbones for low-resolution FGIR: We introduce LRNets, a novel family of CNN architectures tailored for fine-grained recognition under low-resolution constraints. LRNets improve feature extraction directly through architectural modifications to preserve discriminative detail in early stages. Remarkably, LRNets improve accuracy by up to 23\% over prior state-of-the-art (SotA) while requiring up to 20.6x less parameters.

    \item Hardware-aware ViTs trained from scratch: We design ViTFS, a family of Vision Transformers trained entirely from scratch using TGDA, incorporating Batch Normalization for improved hardware compatibility. Notably, ViTFS-Tiny outperforms a pretrained DeiT-B and matches the performance of a ViT-B/16 trained on ImageNet-21k, while using 15.3× fewer parameters.
    
    \item State-of-the-art accuracy with reduced cost: across three fine-grained datasets under diverse settings, TGDA-trained models consistently match or surpass pretrained state-of-the-art methods while significantly reducing inference cost (FLOPs, parameters). TGDA provides an accessible alternative to pretraining, lowering the barrier to entry for research groups and practitioners with limited compute resources.
\end{itemize}

\section{TGDA Framework for Training from Scratch}
\label{sec_method}

\begin{figure*}[!htb]

    \begin{center}
        \includegraphics[width=1.0\linewidth]{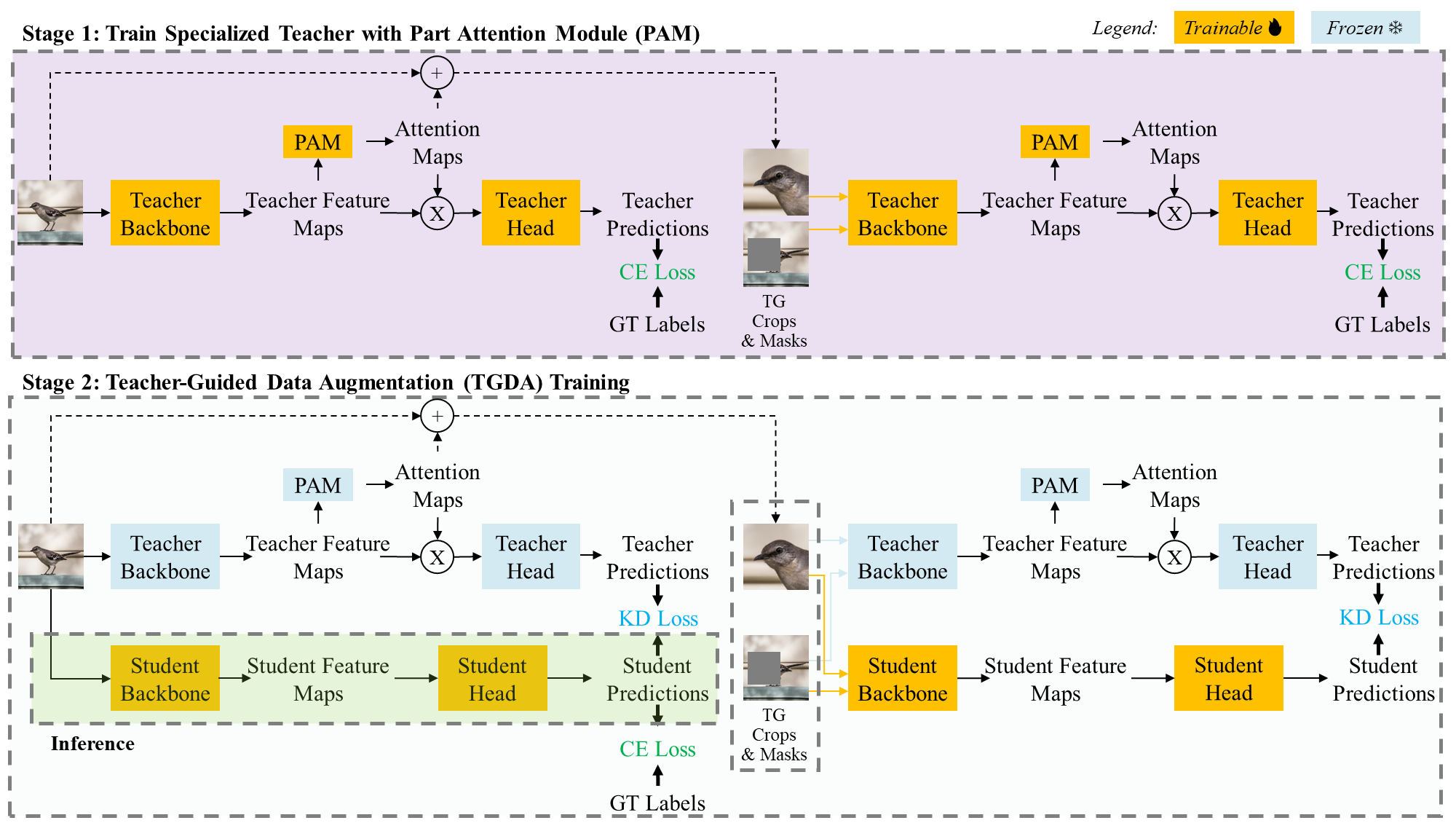}
    \end{center}

    \caption{Overview of the TGDA framework. TGDA is a two-stage training pipeline for fine-grained image recognition from scratch. First, a specialized teacher model is trained with part attention module (PAM) to produce discriminative attention maps. These maps are used to generate diverse, data-aware augmentations via attention cropping and attention dropping. Second, the frozen teacher then supervises a student model using both the original and augmented images through knowledge distillation (KD). The student learns from cross-entropy (CE) loss on original images and KD losses on both original and augmented views, enabling the training of compact, high-performance models without external pretraining.}

    \label{figure_overview_tgda}

\end{figure*}

To address the challenges faced when training fine-grained image recognition models from scratch we propose Teacher-Guided Data Augmentation (TGDA), a two-stage knowledge distillation framework that enables training fine-grained students from scratch using specialized teacher backbones to generate data-aware augmentation for the student. An overview is shown in \cref{figure_overview_tgda}.

\subsection{Stage 1: Specialized Teacher Training with Part Attention Module}

In the first stage, we train a specialized teacher model based on fine-grained-aware architectures such as WS-DAN \cite{hu_see_2019} or CAL \cite{rao_counterfactual_2021}. These models incorporate a Part Attention Module (PAM), a learnable convolutional mechanism that produces diverse attention maps highlighting semantically meaningful object regions. 

The attention maps are combined with intermediate feature maps via bilinear pooling to generate part-aware feature vectors, which are used for classification. These attention maps also serve as the basis for two data-aware augmentations: (1) attention cropping, which zooms into discriminative regions, and (2) attention dropping, which masks those regions to encourage complementary feature learning. This enriched spatial understanding lays the foundation for the student training in Stage 2.

\subsection{Stage 2: Student Training with TGDA}

With the trained teacher frozen, we generate augmentations for each training image and pass both original and augmented samples to the student model. During training, the student receives: 1) cross-entropy (CE) supervision from ground-truth labels (original images only), 2) knowledge distillation (KD) supervision from the teacher’s soft predictions on both original and augmented images.

The student is trained using a combined loss:

$$
\mathcal{L}_{\text{total}} = \alpha \, \mathcal{L}_{\text{CE}}(I_{\text{org}}) + \beta \left[ \mathcal{L}_{\text{KD}}(I_{\text{org}}) + \mathcal{L}_{\text{KD}}(I_{\text{aug}}) \right]
$$

where: $\mathcal{L}_{\text{CE}}$ is cross-entropy loss on the original image, $\mathcal{L}_{\text{KD}}$ is the KL-divergence between the teacher and student outputs for both original and augmented images, and $\alpha$ and $\beta$ control the loss balance.

This training strategy enhances the student’s ability to generalize by exposing it to diverse fine-grained variations and preserving supervision consistency through the teacher.

\section{Task-Specific and Hardware-Aware Backbone Design}

\subsection{LRNet for Low-Resolution FGIR}

To address the challenges of fine-grained recognition under low-resolution (LR) input, we propose LRNet, a task-specific backbone designed to preserve discriminative details often lost in standard architectures. Unlike prior methods, which use off-the-shelf backbones without modification, we explicitly redesign the architecture to better accommodate limited visual resolution.

LRNet extends the standard ResNet design by using five residual blocks instead of four, but modifies the spatial downsampling schedule. Specifically, the initial downsampling rate after the stem is reduced from $4\times$ to $2\times$, and the first residual block preserves the spatial resolution. Downsampling resumes from the second block onward, maintaining the original total reduction factor of $32\times$ by the final layer. This shift delays early spatial resolution loss, allowing the network to retain more fine-grained cues in early representations—critical for distinguishing subtle inter-class differences in LR FGIR.

\subsection{ViTFS for Efficient ViTs from Scratch}

We also extend TGDA to enable training Vision Transformers from scratch (ViTFS), addressing inefficiencies in standard ViT designs for deployment on real-world hardware. One major bottleneck is the use of LayerNorm \cite{ba_layer_2016}, which is poorly supported on devices \cite{noauthor_ethos-u65_nodate} optimized for BatchNorm \cite{ioffe_batch_2015}, leading to runtime and memory inefficiencies. In our measurements on a Galaxy A53 smartphone \cite{noauthor_qualcomm_nodate}, replacing LayerNorm with BatchNorm results in a 2.6× speedup and 19\% lower peak memory usage during inference.

Our ViTFS architecture incorporates a series of hardware-aware and data-efficient design modifications, grouped into three categories:

\vspace{-0.5cm}

\begin{itemize}
    \item \textbf{Inductive Biases for Data Efficiency}
    \begin{itemize}
        \item \textit{Convolutional stem} \cite{xiao_early_2021} instead of a single linear patch embedding, to better capture local structure.
        \item \textit{Sinusoidal 2D positional encoding} instead of learnable embeddings \cite{beyer_better_2022}, avoiding unnecessary parameterization.
        \item \textit{Global average pooling} replaces the CLS token for final classification \cite{beyer_better_2022}, simplifying the architecture.
    \end{itemize}
    
    \item \textbf{Training Stabilization}
    \begin{itemize}
        \item \textit{LayerScale} \cite{cao_training_2022} to stabilize gradients in deep transformers.
        \item \textit{Register tokens} \cite{darcet_vision_2023}, which help preserve intermediate context as shown effective in transformer variants.
    \end{itemize}
    
    \item \textbf{Hardware Efficiency}
    \begin{itemize}
        \item Use of \textit{BatchNorm} instead of LayerNorm throughout the network.
    \end{itemize}
\end{itemize}

\section{Methodology}


We evaluate our approach on three standard fine-grained benchmarks: FGVC-Aircraft (Aircraft) \cite{maji_fine-grained_2013}, Stanford-Cars (Cars) \cite{krause_3d_2013}, and CUB-200-2011 (CUB) \cite{chen_empirical_2021}. We conduct experiments under both high-resolution ($448\times 448$) and low-resolution ($128\times 128$) settings.

All student models are trained for 800 epochs using AdamW, while ResNet-101 teachers are fine-tuned for 50 epochs with SGD based on the CAL recipe \cite{rao_counterfactual_2021}. To further mitigate overfitting when training from scratch we apply generic data augmentation including random cropping, random horizontal flipping, TrivialAugment \cite{muller_trivialaugment_2021}, along with Label Smoothing \cite{szegedy_rethinking_2016} and Stochastic Depth \cite{huang_deep_2016} for regularization. We set the loss weight hyperparameter for TGDA to $\beta = 10$  and the distillation temperature = 7.

We report top-1 accuracy (\%), and evaluate cost using floating-point-operations (FLOPs) and parameter count. We highlight the best results in \textbf{bold} and \underline{underline} the second best.

\section{Results and Discussion}

\subsection{High-Resolution (HR) Fine-Grained Image Recognition}
\label{Fernando_done}

\begin{table}[!htb]
    \centering
    \caption{Accuracy (\%) comparison with state-of-the-art fine-grained methods using image size $448 \times 448$ on ResNet-18 (RN18) and ResNet-34 (RN34).}
    \label{table_tgda_18_34}
    \begin{tabularx}{\linewidth}{VZZZZZZ}
        \toprule
        \multirow{2}{*}{Method} & \multicolumn{2}{c}{Aircraft} & \multicolumn{2}{c}{Cars} & \multicolumn{2}{c}{CUB}  \\
        \cmidrule(lr){2-3} \cmidrule(lr){4-5} \cmidrule(lr){6-7}
         & RN18 & RN34 & RN18 & RN34 & RN18 & RN34 \\
        \midrule

        \multicolumn{7}{l}{\textit{Pre-trained on ImageNet, then fine-tuned on target dataset}} \\
        Baseline & 88.3 & 89.5 & 91.5 & 92.3 & 82.5 & 84.9 \\
        InPS~\cite{zhang_intra-class_2021} & 89.1 & 90.2 & 92.0 & 92.1 & 84.6 & 86.1 \\
        SnapMix~\cite{huang_snapmix_2021} & 89.3 & 90.4 & 92.2 & 93.2 & 83.6 & 86.4 \\
        S3Mix~\cite{zhang_s3mix_2023} & 89.5 & 90.5 & 92.6 & 93.3 & 85.1 & 86.7 \\
        CEKD~\cite{zhang_cekdcross_2022} & \underline{91.3} & \underline{93.7} & \underline{93.9} & \underline{94.6} & \underline{85.2} & \underline{88.3} \\

        \midrule
        \multicolumn{7}{l}{\textit{Trained directly from scratch on target dataset}} \\
        TGDA & \textbf{93.7} & \textbf{94.2} & \textbf{94.9} & \textbf{95.0} & \textbf{87.5} & \textbf{88.4} \\
        \bottomrule
    \end{tabularx}
\end{table}

We compare our proposed method, TGDA, with prior state-of-the-art knowledge distillation approaches for fine-grained classification in \cref{table_tgda_18_34}. TGDA consistently outperforms previous methods across all datasets when using ResNet-18 and ResNet-34, achieving gains of up to 2.4\% compared to previous state-of-the-art, CEKD \cite{zhang_cekdcross_2022}. Notably, our approach does not require the student and teacher to share the same architecture, unlike prior methods. Furthermore, our student models are trained entirely from scratch, whereas previous works rely on ImageNet-pretrained models. As a result, TGDA is exposed to significantly less data, yet still achieves superior performance.

\begin{table}[!htb]
    \centering
    \caption{Performance comparison with state-of-the-art ViT methods (Acc.\%) using image size $448 \times 448$. The table also includes a comparison of the number of parameters (in millions) and the number of images in the training dataset (in thousands).}
    \label{tab_vitfs_448}
    \begin{tabularx}{\linewidth}{Vooozz}
        \toprule
        Method & Aircraft & Cars & CUB & Params.~(M) & Images~(K) \\
        \midrule
        DeiT-T (IN1k PT) & 84.7 & 87.2 & 82.1 & \underline{5.7} & 1,287 \\
        DeiT-S (IN1k PT) & 88.1 & 90.7 & 85.8 & 22.0 & 1,287 \\
        DeiT-B (IN1k PT) & \underline{90.3} & \underline{92.9} & \underline{88.0} & 86.4 & 1,287 \\
        ViT-B (IN1k PT) & 90.0 & 92.5 & \textbf{90.8} & 86.4 & 14,197 \\
        ViTFS-T (TGDA) & \textbf{92.2} & \textbf{94.5} & 86.3 & \textbf{5.6} & \textbf{6} \\
        \bottomrule
    \end{tabularx}
\end{table}

We compare the results of different ViT-based backbones on \cref{tab_vitfs_448}. Our proposed ViTFS trained with the TGDA framework, outperforms state-of-the-art ViT models pretrained on ImageNet-1K on the Aircraft (+1.9\%) and Cars (+1.6\%) datasets, and achieves comparable results on the CUB dataset. Remarkably, ViTFS uses significantly fewer parameters than the top-performing pretrained models and is trained on orders of magnitude fewer images (over $10^3\times$ less), demonstrating its efficiency without compromising performance.

\subsection{Low-Resolution (LR) Fine-Grained Image Recognition}
\label{Fernando_done}

\begin{table}[!htb]
    \centering
    \caption{Accuracy (\%) comparison with state-of-the-art fine-grained methods using image size $128 \times 128$ on ResNet-18 (RN18) and ResNet-34 (RN34). Results marked with '\textbf{*}' correspond to our proposed LRNet backbone, which has reduced computational cost compared to RN18 and RN34, respectively.}
    \label{table_lrnet_18_34}
    \begin{tabularx}{\linewidth}{VZZZZZZ}
        \toprule
        \multirow{2}{*}{Method} & \multicolumn{2}{c}{Aircraft} & \multicolumn{2}{c}{Cars} & \multicolumn{2}{c}{CUB}  \\
        \cmidrule(lr){2-3} \cmidrule(lr){4-5} \cmidrule(lr){6-7}
         & RN18 & RN34 & RN18 & RN34 & RN18 & RN34 \\
        \midrule
        \multicolumn{7}{l}{\textit{Pre-trained on ImageNet, then fine-tuned on target dataset}} \\
        
        Baseline & 50.2 & 52.4 & 60.6 & 64.7 & 59.4 & 62.8 \\
        KD~\cite{hinton_distilling_2015} & 58.7 & 56.0 & 72.1 & 64.8 & 65.0 & 64.9 \\
        CRD~\cite{tian_contrastive_2019} & 53.4 & 56.2 & 60.9 & 65.8 & 61.8 & 66.1 \\
        DKD~\cite{zhao_decoupled_2022} & 58.0 & 60.1 & 75.4 & 79.0 & 71.0 & 68.1 \\
        SnapMix~\cite{huang_snapmix_2021} & 58.5 & 61.8 & 71.0 & 76.8 & 67.9 & 67.1 \\
        DADKD~\cite{zhang_data_2025} & 65.6 & 68.5 & 80.6 & 85.3 & 73.3 & 70.0 \\
        DSSD~\cite{liang_dynamic_2024} & 66.3 & 70.0 & 81.5 & 86.2 & 74.1 & 70.2 \\
        
        \midrule
        \multicolumn{7}{l}{\textit{Trained directly from scratch on target dataset}} \\
        
        TGDA & \underline{87.0} & \underline{85.5} & \underline{91.2} & \underline{91.0} & \underline{77.7} & \underline{78.8} \\
        TGDA+LRNet* & \textbf{89.0} & \textbf{90.7} & \textbf{92.7} & \textbf{93.2} & \textbf{80.5} & \textbf{82.5} \\
        \bottomrule
    \end{tabularx}
\end{table}

Through our TGDA framework, we introduce a novel backbone family called \textbf{LRNet}, specifically designed for low-resolution fine-grained recognition tasks. \cref{table_lrnet_18_34} present a performance comparison between our proposed TGDA + LRNet and prior state-of-the-art methods.

LRNet consistently outperforms existing approaches across various configurations, despite being trained from scratch, unlike most previous methods that rely on ImageNet pretraining. The largest observed performance gap exceeds 20\% in accuracy compared to fine-tuned baselines. Overall, using only the TGDA framework yields a 10\% increase in accuracy on average compared to previous state-of-the-art, DSSD \cite{liang_dynamic_2024}. When further combined with the LRNet backbone, we observe an average performance gain of 13\%, demonstrating both the value of designing a dedicated backbone like LRNet for low-resolution tasks and the efficacy of TGDA in facilitating such architectural innovations.

Furthermore, we remark from \cref{figure_acc_vs_cost_highlights} that these results are obtained with less FLOPs and trainable parameters, while requiring significantly less training data.

\subsection{Ablation on TGDA Differences Compared to Previous Works}

\cref{table_ablation_tgda} highlights the effectiveness of key TGDA design choices. First, applying a student-to-teacher loss as in prior cross-ensemble methods \cite{zhang_cekdcross_2022} yields minimal gains and even harms performance on CUB, confirming that training a teacher via an untrained student is suboptimal. Introducing our Part Attention Module (PAM) alone significantly improves results—especially on CUB demonstrating the superiority of learnable part attention over CAMs. When combined with our two-stage training approach, where the teacher is trained independently before guiding the student, TGDA achieves the best performance across all datasets. This validates that both PAM and two-stage training are critical for effective training from scratch.

\begin{table}[!htb]
    \centering
    \caption{Ablation study on the components of TGDA using image size $128 \times 128$. "PAM" refers to the Part Attention Module, "Two-Stage" denotes the separate training of teacher and student networks, and "S-to-T Loss" corresponds to the cross-ensemble strategy used in prior methods \cite{zhang_cekdcross_2022, liang_dynamic_2024}.}
    \label{table_ablation_tgda}
    \begin{tabularx}{\linewidth}{cccZZZ}
        \toprule
        PAM & Two-Stage & Student-to-Teacher Loss & Aircraft & Cars & CUB \\
        \midrule
        - & - & - & 1.7 & 0.6 & 10.1 \\
        - & - & $\checkmark$ & 2.3 & 0.6 & 1.6 \\
        $\checkmark$ & - & - & 1.0 & 0.6 & 27.1 \\
        $\checkmark$ & $\checkmark$ & - & \textbf{89.0} & \textbf{92.7} & \textbf{80.5} \\
        \bottomrule
    \end{tabularx}
\end{table}

\section{Conclusion}

This paper introduces Teacher-Guided Data Augmentation (TGDA), a novel training framework that enables fine-grained image recognition (FGIR) models to be trained entirely from scratch. By combining data-aware augmentation with weak supervision from a fine-grained-aware teacher, TGDA eliminates the reliance on large-scale pretraining while preserving strong performance.

Through TGDA, we demonstrate that task-specific and hardware-aware architectures such as LRNets for low-resolution FGIR and ViTFS, a family of BN-based Vision Transformers, can be effectively trained from random initialization. Our extensive experiments show that TGDA-trained models not only match or exceed the accuracy of state-of-the-art pretrained counterparts, but do so with significantly reduced inference costs and training footprints. Remarkably, LRNets improve accuracy by up to 23\% over prior state-of-the-art (SotA) while requiring up to 20.6x less parameters. Furthermore, ViTFS Tiny matches the performance of a ViT-B pretrained on ImageNet-21k while using 15.3× fewer trainable parameters.

Together, these results question the necessity of pretraining for FGIR and highlight TGDA as a scalable, deployable, and resource-efficient alternative. We hope TGDA paves the way for more flexible and accessible vision systems, particularly in domains with limited data, compute, or deployment constraints.


\bibliography{references}

\end{document}